\documentclass[runningheads]{llncs}
\usepackage[T1]{fontenc}
\usepackage{graphicx}

\usepackage{amsmath}
\usepackage{amssymb}
\usepackage{booktabs}
\usepackage{listings}
\usepackage{hyperref}
\usepackage{color}
\usepackage{placeins}
\usepackage{float}

\begin{document}
\title{Text-Informed Training Data Filtering for Enhanced Virtual Flow Meter Performance}
\titlerunning{Text-Informed Training Data Filtering}
\author{Kristoffer Hjelmen Nesland\inst{1,2} \and
Bjarne Grimstad\inst{2} \and
Antorweep Chakravorty\inst{1} \and
Dan Sui\inst{1}}
\authorrunning{K. H. Nesland et al.}
\institute{University of Stavanger, Stavanger, Norway \and
Solution Seeker AS, Oslo, Norway\\
\email{krnes5410@uis.no}}
\maketitle
\begin{abstract}
In sectors such as energy, aquaculture and chemical processing, the effectiveness of both mechanistic and data-driven models is frequently limited by poor-quality training data, which obscures underlying physical laws and degrades predictive accuracy. Such quality issues commonly manifest as anomalies in the dataset, ranging from sensor faults to process upsets and incorrect operator inputs. While traditional rule-based filtering can remove a subset of the anomalies, it is unable to leverage contextual information stored as unstructured text, such as engineer comments and operational logs that explain under what conditions the data was collected. We propose a text-informed data filtering method that leverages large language models (LLMs) to integrate unstructured operational data with tabular sensor data. The method consists of two steps: (1) contextual anomaly detection by reasoning over both data modalities, and (2) generation of a binary mask that is applied to filter the training dataset for model development. To investigate the potential benefit, a steady-state soft-sensor regression model, namely a virtual flow meter that estimates well flow rates, is used as the downstream evaluation target. Experiments on a synthetic oil and gas dataset show improved model performance compared to traditional rule-based filtering. Ablation studies show that textual operator comments can provide filtering-relevant signals beyond sensor data alone, with the multimodal approach reducing downstream MAPE from 7.1\,\% to 5.4\,\% relative to sensor-only filtering. This work demonstrates that integrating unstructured operational knowledge with sensor data through LLMs can improve data quality for industrial machine learning applications, offering a scalable alternative to manual data curation.

\keywords{Multimodal data fusion \and Industrial machine learning \and Data-centric AI \and Text-informed data filtering \and Virtual flow meter.}
\end{abstract}
\section{Introduction}\label{sec:introduction}

The attention of researchers and practitioners has gradually shifted from advancing model design to enhancing the quality and quantity of the data~\cite{zha2025}. This data-centric paradigm is particularly relevant in industrial settings, where an increasing amount of sensor data is captured but data quality issues and anomalies are significant due to harsh operational environments~\cite{gaddam2020}. If not addressed, the data quality issues will continue to affect downstream modelling tasks that depend on the data. However, manual quality control of the data is infeasible in many cases due to the large volume of data and the challenge of synthesising information across disparate data streams and modalities~\cite{patel2025}.

\begin{figure}[t]
\centering
\includegraphics[width=0.95\textwidth]{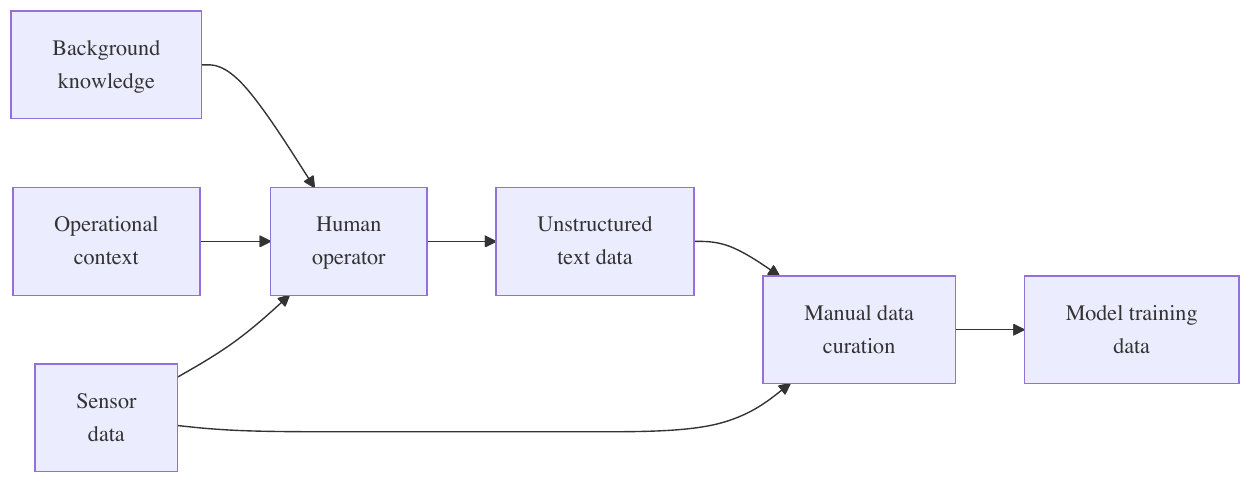}
\caption{Traditional flow of information in industrial data curation. The operator synthesises background domain knowledge, operational context, and sensor readings into unstructured comments. These comments, together with the raw sensor data, then inform a manual curation step that determines which data points enter model training.}\label{fig:traditional-flow}
\end{figure}

Classical anomaly detection methods, as surveyed by Han et al.~\cite{han2022} and consolidated in the PyOD toolbox~\cite{zhao2019}, are unimodal: they operate on sensor data alone. Most existing benchmarks follow the same pattern, with little coverage of settings where relevant quality signals are distributed across multiple modalities~\cite{zha2025}. Yet, to successfully identify anomalous data, sensor readings need to be seen relative to contextual knowledge of the asset. This task involves sifting through unstructured textual data like maintenance logs, shift handovers and technician notes. Consequently, subject matter experts have taken on a role as the bridge between the two data modalities and spent considerable time integrating information across dashboards, PDFs and more (Fig.~\ref{fig:traditional-flow}).

Advances in artificial intelligence (AI) open up new possibilities for bridging the gap between sensor data and unstructured textual data. Today, LLMs can process natural language alongside structured numerical tables within a single context window, making it feasible to reason jointly over operator comments and sensor measurements without human involvement~\cite{hegselmann2023}.

\subsection{Problem Definition}\label{subsec:problem-definition}

Following Chandola et al.~\cite{chandola2009}, an \textit{anomaly} is a pattern in data that does not conform to a well-defined notion of normal behaviour. While ``anomaly'' and ``outlier'' are often used interchangeably in the literature, this paper uses ``anomaly'' throughout. The term is preferred because the deviations considered here stem from identifiable physical events or instrument failures, rather than purely statistical deviation from a distribution. Chandola et al.~\cite{chandola2009} further distinguish between \textit{point anomalies}, which are data instances that deviate from the rest of the data in an absolute statistical sense, and \textit{contextual anomalies}, which are only anomalous given additional context. The anomalies encountered in industrial production data are often contextual in nature and a sensor reading that falls within historical statistical ranges may still represent unreliable data if the well was undergoing an instrument fault at the time of measurement. Detecting such anomalies purely from statistical properties of the sensor readings is therefore insufficient.

The problem addressed in this paper is: given a dataset of paired sensor measurements and operator comments for a set of assets, determine which data points should be classified as anomalies and thereby excluded before model training. We formulate the problem as assigning a binary weight per data point, where one means ``include'' and zero means ``exclude''. A continuous weighting scheme, where anomalous points are down-weighted rather than excluded, was considered but would require determining a suitable degree of down-weighting, which is non-trivial and can depend on the choice of downstream model.

Open industrial datasets such as the Petrobras 3W dataset for well failures~\cite{vargas2019}, the Volve field repository~\cite{equinor2018}, and the ManyWells simulator~\cite{grimstad2026} exist, but none pair sensor measurements with unstructured operator comments and ground-truth anomaly labels. This motivates a synthetic benchmark approach.

\subsection{Contributions}\label{subsec:contributions}

Our contributions are:
\begin{enumerate}
\item A text-informed data filtering method that uses an LLM to jointly reason over tabular sensor measurements and unstructured operator comments, producing a binary data mask.
\item Experimental results on a synthetic oil and gas dataset showing that the multimodal method outperforms classical sensor-based filters on downstream soft-sensor performance, and that unstructured operator comments carry filtering-relevant signals invisible to sensor-based methods alone.
\end{enumerate}

\section{Related Work}\label{sec:related-work}

\subsection{Language Models for Tabular Data and Multimodal Data Fusion}\label{subsec:llm-tabular}

Brown et al.~\cite{brown2020} demonstrate that large language models can perform few-shot learning across a wide range of tasks, and subsequent work has extended this to tabular data classification~\cite{hegselmann2023} and tabular anomaly detection~\cite{tsai2024}. Wu et al.~\cite{wu2026} further demonstrate multimodal in-context anomaly detection, combining structured and unstructured inputs within a single LLM context. However, none of these methods are applied to the problem of data quality filtering for industrial model training, where the goal is to improve downstream predictive performance rather than to classify anomalies as an end in itself.

\subsection{Soft-Sensors and Virtual Flow Metering}\label{subsec:soft-sensors-related}

\textit{Soft-sensor} models use available process measurements to estimate variables that are difficult to measure directly, with regression-based methods surveyed in Souza et al.~\cite{souza2016}. In oil and gas, neural network-based \textit{virtual flow meters} have demonstrated strong performance across diverse well configurations~\cite{sandnes2021}. The predictive quality of such models is sensitive to the training data they are fitted on, yet the effect of systematic data filtering on downstream soft-sensor performance has received little attention. This paper uses a simple soft-sensor model as the downstream evaluation target, making the impact of each filtering strategy directly measurable.

\section{Benchmark Dataset}\label{sec:benchmark}

\subsection{Dataset Implementation}\label{subsec:dataset-implementation}

We construct a synthetic benchmark, built on ManyWells~\cite{grimstad2026}, a steady-state multiphase flow simulator for oil and gas wells with varying reservoir properties, wellbore geometries and fluid compositions. The simulator uses a one-dimensional drift-flux model and has been validated against real production data from more than 300 wells. A \textit{well test} is a separator flow test in which the well is temporarily routed through a dedicated test separator to measure the volumetric flow rates of oil, gas and water, with mean values over a stable test interval reported as a single observation. Each ManyWells simulation thus represents one such well-test data point, described by measurements of pressures, temperatures, choke position, phase flow rates and more. Reservoir pressure and flow composition are assumed constant per well, so the benchmark does not model temporal effects such as reservoir depletion.

\subsubsection{Data Generation.}
Let $J$ denote the set of all wells in the benchmark. For each well $j \in J$, the simulator produces $N_j$ data points by sampling operating conditions (choke opening, gas lift rate) from the well's feasible domain and solving the steady-state flow equations. Each data point is described by a sensor measurement vector $\mathbf{x}_{ij} \in \mathbb{R}^d$ and an optional textual comment $t_{ij} \in \mathcal{T} \cup \{\varnothing\}$, where $i = 1, \ldots, N_j$ indexes the data points within well $j$, $\mathcal{T}$ is the set of all non-empty strings and $\varnothing$ denotes the absence of a comment. A subset of data points is then designated as anomalous through controlled perturbations of either the simulation parameters or the resulting sensor readings. The ground-truth weight is $w_{ij} = 0$ whenever any perturbation is applied and $w_{ij} = 1$ otherwise.

The wells are partitioned into three roles:
\begin{itemize}
\item \textit{Reference wells} ($J_{\text{ref}} \subset J$) with disclosed ground-truth weights, used as in-context examples for the LLM.
\item \textit{Training wells} ($J_{\text{dev}} \subset J$) with undisclosed weights, on which all filtering methods are applied.
\item \textit{Evaluation splits}: each training well also has a held-out evaluation set generated under nominal conditions (no anomalies), used solely for measuring downstream model performance.
\end{itemize}

\subsubsection{Textual Comments.}
Each injected anomaly carries a textual comment $t_{ij}$ inspired by real-world operator logs from Norwegian offshore operations. Comments are written in a mixture of Norwegian and English, reflecting actual communication practices, and vary in specificity from precise diagnoses (e.g.\ ``Injecting MEG to dissolve hydrate plug after unplanned shutdown'') to vague observations (e.g.\ ``Potential for scaling''). Some anomalous data points carry an empty comment ($t_{ij} = \varnothing$), mimicking the common case where anomalies go unreported. Conversely, some nominal data points carry positive or neutral comments (e.g.\ ``Test went according to plan''), as occurs in practice when operators note routine events.

\subsection{Dataset Details}\label{subsec:dataset-details}

\subsubsection{Sensor Features.}
Each sensor measurement vector $x_{ij}$ contains $d=4$ features: Bottomhole pressure ($P_{BH}$, bar), Wellhead pressure ($P_{WH}$, bar), Volumetric liquid flow rate ($Q$, Sm$^3$/h) and Wellbore length ($L$, m). This is only a small subset of the measurements provided by ManyWells. For a thorough analysis of the ranges of the features, we refer to the ManyWells paper~\cite{grimstad2026}.

\subsubsection{Anomaly Categories and Injection Mechanisms.}
Anomalies are grouped into six categories that reflect common operational issues in oil and gas production (Table~\ref{tab:anomaly-groups}). The categories are inspired by Theyab~\cite{theyab2018}.

\begin{table}[t]
\caption{Anomaly groups.}\label{tab:anomaly-groups}
\centering
\begin{tabular}{lll}
\toprule
Category & Perturbation type & Physical mechanism \\
\midrule
Scaling & Well parameter ($f_D$, $D$) & Mineral deposits restricting flow \\
Hydrate formation & Well parameter ($f_D$, $D$) & Ice-like plugs in the wellbore \\
Liquid loading & Well parameter ($\boldsymbol{\alpha}$) & Inability to lift liquids \\
Emulsion & Well parameter ($f_D$) & Oil-water emulsions increasing friction \\
Wax deposition & Well parameter ($f_D$) & Wax build-up increasing friction \\
Data quality & Simulated parameter & Sensor faults and communication failures \\
\bottomrule
\end{tabular}
\end{table}

The well parameters perturbed are the wellbore Darcy friction factor $f_D$, the wellbore internal diameter $D$, and the phase composition vector $\boldsymbol{\alpha}$ giving the gas, oil and water fractions. Two perturbation mechanisms are used. \textit{Well parameter perturbations} modify physical parameters before simulation, producing data points that are internally consistent but reflect abnormal well conditions such as flow restrictions from scaling, wax or emulsions, physical obstructions from hydrate plugs, or liquid loading. \textit{Simulated parameter perturbations} modify sensor readings after simulation to model instrument faults such as transmitter drift or flatlined meters.

\subsubsection{Dataset Sizes.}
We utilise 13 well configurations spanning a range of fluid compositions, splitting them into three reference wells ($J_{\text{ref}} = \{51, 776, 1234\}$) used for in-context examples, and ten training wells\newline
($J_{\text{dev}} = \{289, 406, 641, 715, 745, 1100, 1580, 1773, 1820, 1846\}$). The indexes refer to existing configurations in the ManyWells dataset. Training set sizes vary from 10 to 60 data points per well, with 50 evaluation points per well. In total, the benchmark contains 80 reference, 311 training and 500 evaluation data points. Each training well has, on average, 10 injected anomalies placed at random positions, alongside non-anomalous data points. The anomaly rate across wells varies from roughly 30\,\% to 60\,\% of the training data.

The distribution of the textual data is designed to be broadly representative of the Norwegian offshore asset that motivated the benchmark design, though comment coverage and specificity vary across assets depending on operational culture and documentation practices. Of anomalous points, 55.4\,\% carry a comment describing the fault condition and the remaining 44.6\,\% go unreported, reflecting the case where anomalies are not logged. Of nominal points, 92.9\,\% carry no comment, with the remaining 7.1\,\% receiving a routine note confirming normal operation. For simplicity, anomalous points never receive routine comments, and nominal points never receive anomaly-indicating ones.

\section{Method}\label{sec:method}

\subsection{Problem Formulation}\label{subsec:problem-formulation}

For each well $j \in J$, the dataset $\mathcal{D}_j = \{(\mathbf{x}_{ij}, t_{ij})\}_{i=1}^{N_j}$ collects $N_j$ data points as defined in the benchmark description above.

The goal is to determine a binary weight vector $\mathbf{w}_j = (w_{1j}, \ldots, w_{N_j j}) \in \{0,1\}^{N_j}$ for each well, where $w_{ij} = 1$ indicates that data point $i$ should be retained for model training and $w_{ij} = 0$ indicates that it should be excluded as anomalous. In total, a number of $N = \sum_{j \in J_{\text{dev}}} N_j$ weights need to be determined. A weighting function is a function $\mathcal{F}$ that maps the dataset to a weight vector:

\begin{equation}\label{eq:filtering-function}
\mathcal{F}: \mathcal{D} \mapsto \mathbf{w}
\end{equation}

The weight vector is subsequently used in a weighted regression objective when fitting a downstream model. A good filtering method should exclude data points that are anomalous (due to physical faults or sensor errors) while retaining data that reflects normal well behaviour, thereby improving the predictive accuracy of the fitted model on the nominal evaluation data.

\subsection{LLM-Based Multimodal Anomaly Detection}\label{subsec:llm-multimodal}

The proposed anomaly detection method uses an LLM to integrate both sensor measurements $\mathbf{x}_{ij}$ and textual comments $t_{ij}$ into a single anomaly-classification decision per data point. This multimodal approach is contrasted with traditional statistical baselines that operate on sensor measurements alone.

The LLM is given a structured prompt containing: (1) a task description specifying that the goal is to assign binary weights to each sample, (2) a feature dictionary explaining the physical meaning and units of each variable, (3) $|J_{\text{ref}}|$ reference datasets with disclosed weights as in-context examples and (4) the target training dataset with unknown weights. Each dataset is presented as a JSON array of records, with each record containing an input object (sensor values and comment) and an output object (the weight to be predicted). The LLM returns a structured JSON response containing one decision per data point, each consisting of a binary weight and a brief natural-language reasoning.

The system instructions include a feature dictionary (component 2) because the JSON field names are domain-specific abbreviations whose physical meaning and units are not self-evident. Providing explicit column semantics is standard practice when serialising tabular data for LLMs, enabling the model to draw on pre-trained domain knowledge~\cite{hegselmann2023,tsai2024}. The reference datasets in component (3) follow the few-shot in-context learning paradigm of Brown et al.~\cite{brown2020}.

Formally, let $\mathcal{S}$ denote the system instructions (task description and feature dictionary), and let $\mathcal{R} = \{\mathcal{D}_j\}_{j \in J_{\text{ref}}}$ denote the set of reference datasets with known weights. The LLM filtering function is applied independently to each training well $j \in J_{\text{dev}}$:

\begin{equation}\label{eq:llm-filtering}
\mathcal{F}_{\text{LLM}}: (\mathcal{S},\, \mathcal{R},\, \mathcal{D}_j) \mapsto \mathbf{w}_j, \quad j \in J_{\text{dev}}
\end{equation}

where $\mathcal{D}_j$ is the target training dataset for well $j$ (with weights unknown), as defined above, and the LLM performs in-context learning from $\mathcal{R}$ without any gradient-based training. A single sample $(\mathbf{x}_{ij}, t_{ij}, w_{ij})$ from one of the reference wells is shown in Fig.~\ref{fig:json-sample}.

\begin{figure}[H]
\centering
\begin{lstlisting}
{
  "INPUT": {
    "WELL_IDX": 51, "PBH": 261.8, "PWH": 71.9, "QLIQ": 89.5, "L": 3390.6,
    "COMMENT": "WHP transmitter drifting. Reading static despite flow variations."
  },
  "OUTPUT": { "WEIGHT": 0 }
}
\end{lstlisting}
\caption{A sample in the utilised JSON format.}\label{fig:json-sample}
\end{figure}

\subsection{Baseline Methods}\label{subsec:baseline-methods}

The baselines operate on liquid flow rate ($Q$) and pressure differential ($P_{\text{BH},i} - P_{\text{WH},i}$). Wellbore length ($L$) is not included as it is a constant per well and contributes no within-well discriminating signal. This is in contrast with the LLM method where the model can potentially gain some insights from comparing the wellbore length with the reference wells.

Two classical anomaly-detection methods are applied:
\begin{enumerate}
\item \textit{Tukey's fences}~\cite{tukey1977}: compute the interquartile range ($IQR = Q_3 - Q_1$) for each axis and flag points outside $[Q_1 - 1.5 \cdot \text{IQR},\; Q_3 + 1.5 \cdot \text{IQR}]$ on either axis as anomalous ($w_{ij} = 0$). Here, $Q_1$ is the first quartile and $Q_3$ is the third quartile.
\item \textit{Robust Mahalanobis}~\cite{hubert2018}: fit a Minimum Covariance Determinant (MCD) estimator, with support fraction of 0.85, to the data and compute the squared Mahalanobis distance $d_{ij}^2$ for each point. Points exceeding the $\chi^2_p(0.95)$ quantile are flagged ($w_{ij} = 0$), where $p$ is the number of features.
\end{enumerate}

Two additional reference methods are included:
\begin{enumerate}
\item Uniform: $w_{ij} = 1$ for all $i \in \{1, \ldots, N_j\}$, $j \in J_{\text{dev}}$ (no filtering), establishing a lower bound on performance.
\item Oracle: $w_{ij} = w_{ij}^{*}$, where $w_{ij}^{*}$ denotes the ground-truth weight from the simulator, establishing an upper bound.
\end{enumerate}

\subsection{Ablation and Prompt Augmentation}\label{subsec:ablation-augmentation}

To study the contribution of each modality, two more input views are defined. The \textit{text-only view} strips all sensor measurements, providing only the engineer comment for each row. The \textit{sensor-only view} removes the comment field.

To investigate whether explicitly presenting statistical context improves LLM performance, we study two augmented variants that provide pre-computed anomaly statistics in the prompt. The first one augments the prompt with a binary flag for each sample, calculated with the Tukey algorithm. The second one injects the Mahalanobis distance for each data point into the prompt.

\section{Experimental Setup}\label{sec:experimental-setup}

\subsection{Research Questions and Metrics}\label{subsec:research-questions}

Our research questions are:
\begin{enumerate}
\item \textbf{Multimodal LLM vs.\ classical}: Does multimodal LLM filtering, combining sensor data and operator comments, outperform classical rule-based methods (Tukey, Mahalanobis) on downstream soft-sensor performance?
\item \textbf{Value of textual information in LLM filtering}: Do operator comments provide filtering-relevant signals beyond what sensor data alone captures when used as LLM input?
\item \textbf{Classical anomaly scores in the prompt}: Does adding pre-computed classical anomaly scores to the multimodal LLM prompt outperform multimodal-only LLM filtering?
\end{enumerate}

All research questions are evaluated on downstream soft-sensor quality: mean absolute percentage error (MAPE), mean squared error (MSE), and fraction of predictions within 10\,\% of the true value (W10). ``Outperforms'' means strictly better on two of these three reported metrics. Table~\ref{tab:method-rq} summarises the role of each method for each research question.

\begin{minipage}{\linewidth}
MAPE, MSE, and W10 are calculated as:
\begin{itemize}
\item \textbf{MAPE}: $\frac{100}{n}\sum_{i=1}^{n} \left|\frac{Q_{i} - \hat{Q}_{i}}{Q_{i}}\right|$.
\item \textbf{MSE}: $\frac{1}{n}\sum_{i=1}^{n} (Q_{i} - \hat{Q}_{i})^2$.
\item \textbf{W10}: $\frac{100}{n}\sum_{i=1}^{n} \mathbf{1}\!\left[\left|\frac{Q_{i} - \hat{Q}_{i}}{Q_{i}}\right| \leq 0.10\right]$, where $\mathbf{1}[\cdot]$ equals 1 if the condition holds and 0 otherwise.
\end{itemize}

Here, $n$ is the number of evaluation points.
\end{minipage}

\begin{table}[t]
\caption{Role of each method per research question. B = baseline, $\bullet$ = tested, abl. = ablation, --- = reference bound.}\label{tab:method-rq}
\centering
\begin{tabular}{lccc}
\toprule
Method & RQ1 & RQ2 & RQ3 \\
\midrule
Uniform (no filter) & --- & --- & --- \\
Tukey & B & & \\
Mahalanobis & B & & \\
LLM (multimodal) & $\bullet$ & $\bullet$ & B \\
LLM (text only) & & abl. & \\
LLM (sensor only) & & B & \\
LLM + Tukey & & & $\bullet$ \\
LLM + Mahalanobis & & & $\bullet$ \\
Oracle (ground truth) & --- & --- & --- \\
\bottomrule
\end{tabular}
\end{table}

\subsection{LLM Inference}\label{subsec:llm-inference}

The LLM filtering function $\mathcal{F}_{\text{LLM}}$ is instantiated with Anthropic's Claude Sonnet 4.6~\cite{anthropic2026}, using structured output via JSON schema enforcement. This ensures that the model returns exactly $N_j$ decisions per well in a parseable format. Each decision includes a binary weight and a one-sentence reasoning string. Because LLM inference is stochastic, each LLM method is run $R = 3$ independent times per well. Downstream metrics are computed by pooling the prediction errors from all runs. The repeated runs allow for a direct analysis of filtering consistency across runs, presented in the stability analysis in Section~\ref{subsec:agreement-consistency}.

\subsection{Soft-Sensor Model}\label{subsec:soft-sensor-model}

To evaluate the downstream impact of each filtering method, a simple mechanistic soft-sensor model is fitted per well, $j$. The model predicts the liquid (oil and water) flow rate $Q$ from the pressure differential $\Delta P_i = P_{\text{BH},i} - P_{\text{WH},i}$ and the vertical wellbore length $L_j$. \textit{Net driving pressure} is the differential minus hydrostatic head, truncated at zero:

\begin{equation}\label{eq:net-pressure}
\Delta P_{\text{net},i} = \max\!\left(\Delta P_i - \frac{\rho \, g \, L_j}{C_F},\, 0\right)
\end{equation}

\begin{equation}\label{eq:soft-sensor}
\hat{Q}_i = c \,\sqrt{\Delta P_{\text{net},i}}
\end{equation}

Here $\Delta P_i$ is expressed in bar, $c$ is a flow coefficient that lumps geometry and friction into a single constant, and $\rho$ is an effective fluid density (kg/m$^3$) that absorbs multiphase effects in a single scalar. $g$ is standard gravitational acceleration (m/s$^2$), and $C_F = 10^5$ converts hydrostatic head from pascals to bar (Pa/bar). With $\Delta P_{\text{net},i}$ as net driving pressure, a steady one-dimensional balance in which frictional losses scale with the square of a representative velocity implies a quadratic relation between flow rate and net pressure drop, $\Delta P_{\text{net}} \propto Q^2$, and hence $Q \propto \sqrt{\Delta P_{\text{net}}}$. This is the same structure as a Darcy--Weisbach-type loss in pipe flow~\cite{brown2002}. The parameters $c$ and $\rho$ are fitted per well by minimising the weighted mean squared error, where we drop the $j$ subscript since we optimise per well:

\begin{equation}\label{eq:loss}
c^*,\, \rho^* = \arg\min_{c,\,\rho} \; \frac{1}{\sum_{i=1}^{N_j} w_i}\sum_{i=1}^{N_j} w_i \left(Q_{i} - \hat{Q}_{i}\right)^2
\end{equation}

\begin{figure}[t]
\centering
\includegraphics[width=0.85\textwidth]{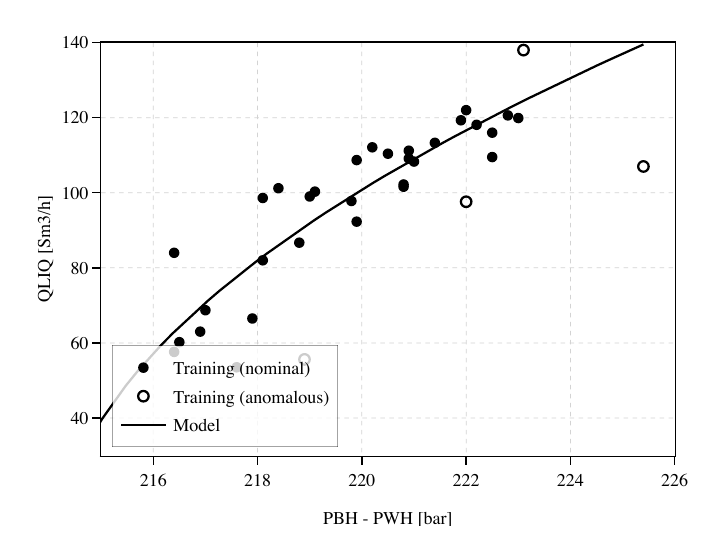}
\caption{Example of a dataset for a single well and the model fitted to the nominal data points.}\label{fig:single-well-model}
\end{figure}

The effective density $\rho$ is found via grid search, while $c$ is solved in closed form for each candidate $\rho$. The model is intentionally simple: the focus of this study is not the model architecture but the effect of different filtering strategies on downstream predictive performance. A more expressive model such as the one proposed by Sandnes et al.~\cite{sandnes2021} would draw on a higher-dimensional feature space, making it difficult to visualise the relationship between training data and model fit in two dimensions. By restricting the model to a single varying predictor per well ($\Delta P$), the relationship between the retained training data and the fitted model can be visualised in two dimensions, keeping the results directly human-interpretable, as illustrated in Fig.~\ref{fig:single-well-model}.

\FloatBarrier

\section{Results and Discussion}\label{sec:results}

\subsection{Downstream Soft-Sensor Performance}\label{subsec:downstream-performance}

All methods share the same model architecture and evaluation set, the only difference is which samples are classified as anomaly and excluded from training. ``Uniform'' retains every sample, including injected anomalies. ``Oracle'' correctly classifies all anomalies. The evaluation set contains no anomalies, ensuring that differences in performance reflect the filtering strategy alone.

Table~\ref{tab:results} reports downstream metrics on the held-out evaluation sets: MSE, MAPE and W10. Predictions are pooled across wells and, for LLM variants, across the $R = 3$ independent classification runs. The 95\,\% confidence intervals are computed by a percentile bootstrap with 10,000 resamples over the pooled evaluation points, taking the 2.5th and 97.5th percentiles as bounds.

\begin{figure}[t]
\centering
\includegraphics[width=0.85\textwidth]{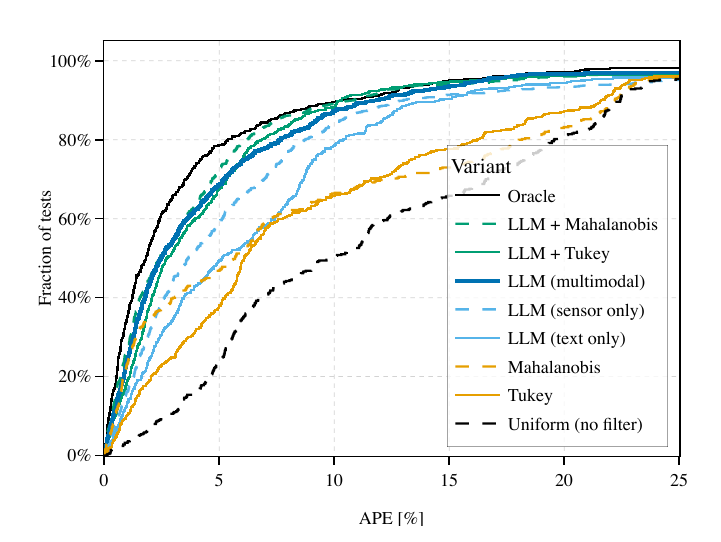}
\caption{Results presented as a cumulative performance plot.}\label{fig:cumulative}
\end{figure}

The role of each method per research question is summarised in Table~\ref{tab:method-rq}. Fig.~\ref{fig:cumulative} presents the results as a \textit{cumulative performance plot}. These are information-dense plots that, given a performance threshold (x-axis), show the percentage of the total evaluation samples (y-axis) where the model has a better or equal performance than the threshold. We address each research question in turn.

\begin{table}[t]
\caption{MSE, MAPE and W10, pooled across wells and classification runs (95\,\% confidence intervals).}\label{tab:results}
\centering
\begin{tabular}{lrrr}
\toprule
Method & MSE ((Sm$^3$/h)$^2$) & MAPE (\%) & W10 (\%) \\
\midrule
Uniform (no filter) & 148 [123, 176] & 12.5 [11.5, 13.6] & 51 [46, 55] \\
\addlinespace
Tukey & 112 [86, 144] & 10.0 [8.9, 11.2] & 66 [62, 70] \\
Mahalanobis & 123 [98, 151] & 9.4 [8.3, 10.6] & 66 [62, 71] \\
\addlinespace
LLM (multimodal) & 53 [44, 64] & 5.4 [4.9, 5.9] & 87 [85, 89] \\
LLM (text only) & 86 [71, 103] & 8.0 [7.4, 8.7] & 78 [76, 81] \\
LLM (sensor only) & 88 [74, 103] & 7.1 [6.5, 7.8] & 84 [82, 86] \\
\addlinespace
LLM + Tukey & 64 [49, 82] & 5.7 [5.1, 6.4] & 88 [87, 90] \\
LLM + Mahalanobis & 47 [36, 59] & 5.1 [4.6, 5.6] & 89 [87, 91] \\
\addlinespace
Oracle (ground truth) & 36 [26, 48] & 4.1 [3.5, 4.8] & 90 [87, 92] \\
\bottomrule
\end{tabular}
\end{table}

Without filtering, Uniform (no filter) performs worst on all reported metrics (MSE 148, MAPE 12.5\,\%, W10 51\,\%). Among the classical filters, the best method depends on the metric: Mahalanobis achieves the lowest MAPE (9.4\,\%), while Tukey achieves the lowest MSE (112~(Sm$^3$/h)$^2$), both reach W10 of about 66\,\%. LLM (multimodal) improves on the best classical result in each case (MSE 53, MAPE 5.4\,\%, W10 87\,\%). The confidence intervals for multimodal LLM filtering do not overlap those of Mahalanobis or Tukey on any of the metrics (Table~\ref{tab:results}), and Fig.~\ref{fig:cumulative} shows the same ordering across the full error distribution. \textbf{RQ1:} multimodal LLM filtering outperforms both Tukey and Mahalanobis on MSE, MAPE, and W10 under the criterion defined in Section~\ref{sec:experimental-setup}.

The key comparison for RQ2 is LLM (multimodal) against LLM (sensor only), which isolates the contribution of the textual modality while holding sensor input constant. Adding operator comments to the sensor-based prompt improves performance on all three metrics: MAPE drops from 7.1\,\% to 5.4\,\% (1.7 percentage points), MSE drops from 88 to 53, and W10 rises from 84\,\% to 87\,\%. The confidence intervals on MAPE do not overlap (sensor only: [6.5, 7.8] vs.\ multimodal: [4.9, 5.9]), indicating a statistically meaningful difference. The same goes for MSE and W10. \textbf{RQ2:} textual operator comments carry filtering-relevant information that is not captured by sensor data alone, and the LLM can leverage this when both modalities are presented jointly.

Augmenting the multimodal LLM prompt with robust Mahalanobis distance improves downstream performance relative to LLM (multimodal) alone. LLM + Mahalanobis performs best among all non-Oracle methods (MSE 47, MAPE 5.1\,\%, W10 89\,\%), improving on multimodal-only by 6 MSE units, 0.3 percentage points of MAPE, and 2 percentage points of W10. The augmentation of the prompt with pre-computed flags from Tukey's fences does not provide the same performance jump and has a negative effect on MSE and MAPE, but a positive effect on W10. \textbf{RQ3:} adding classical anomaly scores to the prompt can improve downstream metrics relative to multimodal LLM filtering alone, but the effect is not universally positive and depends on the selected method.

\subsection{Cross-Method Agreement and Consistency of LLM Filtering}\label{subsec:agreement-consistency}

Beyond downstream soft-sensor metrics, we examine whether the LLM-based filtering methods agree on which data points to classify as anomaly, and how consistently each method reproduces its own decisions across independent runs. To quantify this, we compare the binary weight vectors produced by each LLM method across all training wells ($J_{\text{dev}}$). Because LLM-based methods are stochastic, each LLM method $\mathcal{F}_a$ is executed $R = 3$ independent times, yielding weight vectors $\mathbf{w}_j^{(a,r)} = \mathcal{F}_a^{(r)}(\mathcal{D}_j)$ for $r \in \{1, 2, 3\}$.

We define the pairwise percentage of matching decisions between run $r_a$ of method $a$ and run $r_b$ of method $b$ as

\begin{equation}\label{eq:match-pair}
\mathrm{Match}(a, r_a,\, b, r_b)
= \frac{100}{\sum_{j \in J_{\text{dev}}} N_j}\sum_{j \in J_{\text{dev}}}\left|\left\{\, i \in \{1,\ldots,N_j\} : w_{ij}^{(a,r_a)} = w_{ij}^{(b,r_b)} \,\right\}\right|.
\end{equation}

For each pair of methods $(a, b)$, we report $\mathrm{Match}(a, b)$, the average over all valid run combinations:

\begin{equation}\label{eq:match-avg}
\mathrm{Match}(a, b) = \frac{1}{|P_{ab}|} \sum_{(r_a,\, r_b)\,\in\, P_{ab}} \mathrm{Match}(a, r_a,\, b, r_b),
\end{equation}

where $P_{ab}$ is the set of valid run pairs: $\{(r_a, r_b) : r_a \neq r_b\}$ when $a = b$, and $\{1,\ldots,R_a\} \times \{1,\ldots,R_b\}$ when $a \neq b$.

\begin{table}[t]
\caption{Run-to-run consistency: $\mathrm{Match}(a,a)$ across all training wells (percent matching decisions).}\label{tab:match-run}
\centering
\begin{tabular}{lr}
\toprule
Method & Match (\%) \\
\midrule
LLM (multimodal) & 96.6 \\
LLM (text only) & 100.0 \\
LLM (sensor only) & 92.7 \\
LLM + Tukey & 96.4 \\
LLM + Mahalanobis & 97.6 \\
\bottomrule
\end{tabular}
\end{table}

We first examine run-to-run consistency ($a = b$) presented in Table~\ref{tab:match-run}. LLM (text only) is perfectly stable across all three runs (100\,\% match), whereas LLM (sensor only) agrees with itself on 92.7\,\% of decisions. This gap suggests that sensor-based prompts leave more room for stochastic variation, likely because numeric patterns near the exclusion threshold are harder to resolve consistently than comment-based cues. Adding pre-computed Mahalanobis distances to the prompt increases consistency to 97.6\,\%, up from 96.6\,\% for LLM (multimodal), indicating that classical anomaly scores can stabilise LLM filtering decisions. However, LLM + Tukey is marginally lower at 96.4\,\%.

For the cross-method agreement ($a \neq b$), the largest disagreement appears between the two unimodal baselines: LLM (sensor only) and LLM (text only) match on only 74.1\,\% of decisions. This confirms that the two modalities identify largely different subsets of anomalous points, supporting the view that operator comments encode filtering-relevant context not visible in sensor readings alone (RQ2). LLM (multimodal) agrees with neither unimodal method fully (84.9\,\% vs.\ sensor only, 86.5\,\% vs.\ text only), consistent with joint reasoning over both inputs rather than deferring to a single modality.

\subsection{Explainability of LLM Filtering}\label{subsec:explainability}

\begin{figure}[t]
\centering
\includegraphics[width=0.85\textwidth]{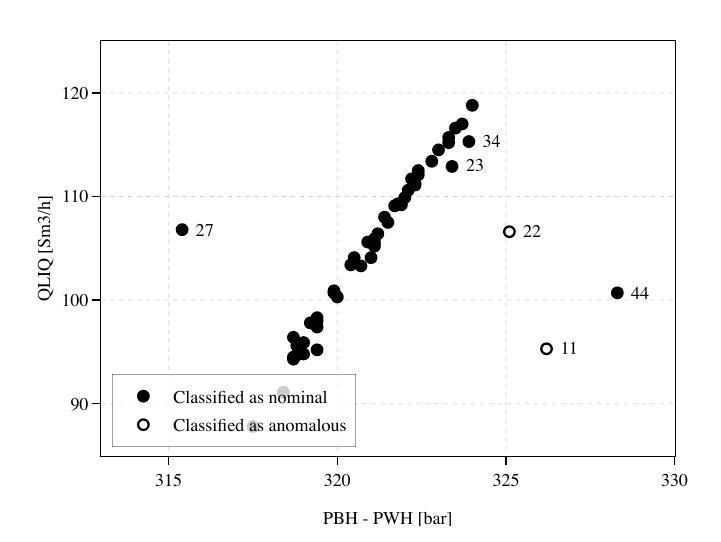}
\caption{Data points of the well with index 745 processed by the LLM (multimodal) method. Data points of particular interest are numbered.}\label{fig:well-745-scatter}
\end{figure}

Unlike Tukey's fences or Mahalanobis distance, which apply transparent arithmetic rules, the LLM methods are fundamentally black-box classifiers. To obtain some insight into the decisions, each LLM call is prompted to produce a one-sentence natural-language justification alongside every binary weight assignment. We illustrate this in Fig.~\ref{fig:well-745-scatter} with data points from well 745 processed by LLM (multimodal). The examples show that the model weighs text against sensor readings rather than just pattern-matching on either.

The model's reasoning for the numbered data points is:
\begin{itemize}
\item 11: ``Comment indicates `Masse emulsjoner' (lots of emulsions), which as seen in reference datasets leads to unreliable measurements.''
\item 22: ``Comment indicates `Emulsjoner i flowback' (emulsions in flowback), which compromises measurement reliability per reference datasets.''
\item 23: ``Comment mentions only minor easily-soluble emulsions (`Kun litt lettl{\o}selige emulsjoner'), not severe enough to exclude.''
\item 27: ``PBH (327.9) is slightly lower than typical but QLIQ is consistent; no anomalous comment.''
\item 34: ``Normal data point with values consistent with the main cluster.''
\item 44: ``PWH of 7.8 is slightly low but PBH and QLIQ are consistent with the cluster; acceptable data point.''
\end{itemize}

Points 11 and 22 together show how operator comments can drive classification independent of proximity to the cluster. Point 11 lies far from the cluster and contains a comment indicating severe emulsion, so it is excluded. As seen in Fig.~\ref{fig:well-745-scatter}, point 22 is considerably closer to the cluster but is excluded nonetheless, because its comment flags emulsions in flowback, a condition the model treats as compromising based on the reference data. The contrasting cases are points 27 and 44. Both points sit at a significant distance from the cluster (Fig.~\ref{fig:well-745-scatter}), and the model acknowledges that the pressure values are somewhat off, but with no anomalous comment and flow rates within the expected range, neither is classified as an anomaly. Point 34 has values that are consistent with the main cluster and no comment. Like point 34, point 23 is also close to the main cluster and is not classified as an anomaly despite its comment hinting at emulsion issues. In sum, the six points of Fig.~\ref{fig:well-745-scatter} show that the model integrates the text and sensor modalities non-trivially. The outcome is not determined by text alone. For instance, point 23 is accepted despite an emulsion hint. Similarly, sensor distance is not the sole deciding factor, as point 11 is classified as an anomaly while point 44 is not, despite both having similar distances from the cluster.

\section{Conclusion}\label{sec:conclusion}

This paper addressed data quality issues in industrial datasets by proposing a multimodal LLM method that jointly reasons over sensor data and unstructured operator comments to filter anomalous training samples. On the downstream regression task, the multimodal method outperforms the sensor-only method by 1.7 percentage points and it is shown that the textual data adds signal beyond the LLM only assessing the sensor data. More broadly, this points to operator comments as an underused data-centric resource: a modality already recorded in industrial workflows that can be folded into model development without the manual curation it currently demands.

The results reported in this paper are based on a synthetic benchmark of 13 wells, and performance on real-world data may deviate. The informativeness of comments is a design choice calibrated to a real asset, and the magnitude of the text benefit depends on it. Further, the study of the model's ``reasoning'' traces provides only a limited insight into the behaviour of inherently opaque LLM methods. Beyond interpretability, LLM-based approaches introduce practical overhead relative to classical rule-based algorithms: inference latency is higher, behaviour is non-deterministic, and reliance on proprietary models raises concerns about data privacy and service availability. These factors may limit applicability in latency-sensitive or security-constrained operational settings. All LLM-based experiments use a single model (Claude Sonnet 4.6). Filtering decisions and downstream metrics may differ for other models.

\subsection{Future Work}\label{subsec:future-work}

Real-world data would help assess how well the results generalise and how sensitive they are to reference dataset size and composition. For industrial applicability, treating time-dependent phenomena remains an open challenge, particularly in determining how long textual information stays valid for anomaly assessment. Regarding explainability, more work is needed to understand the relative importance of sensor data and text.

\begin{credits}
\subsubsection{Reproducibility}

The code for generating the benchmark dataset and reproducing the experiments is publicly available at\newline
\url{https://github.com/krNesland/nesland2026-text-informed-filtering}.

\subsubsection{Generative AI Use}

Generative AI tools were utilised in the work leading up to this paper. Google Gemini was partly used to suggest relevant literature. Anthropic Sonnet 4.6 was used to develop some of the mathematical notation and to increase the readability of the paper. The authors have reviewed and edited all AI-generated content and take full responsibility for the final content.

\subsubsection{\discintname}

K. H. Nesland and B. Grimstad are employed by Solution Seeker AS and A. Chakravorty and D. Sui are at the University of Stavanger. The underlying PhD project is industrially funded (Research Council of Norway, Solution Seeker AS and industry partner). Solution Seeker develops production-optimisation software related to the methods discussed and ManyWells, used for the synthetic benchmark, originates from that research environment. No other competing financial interests are declared.
\end{credits}

\clearpage
\bibliographystyle{splncs04}
\bibliography{references}

@misc{anthropic2026,
  author       = {{Anthropic}},
  title        = {Introducing {Sonnet} 4.6},
  year         = {2026},
  howpublished = {\url{https://www.anthropic.com/news/claude-sonnet-4-6}}
}

@inproceedings{brown2002,
  author    = {Brown, Glenn O.},
  title     = {The History of the {Darcy-Weisbach} Equation for Pipe Flow Resistance},
  booktitle = {Environmental and Water Resources History},
  pages     = {34--43},
  year      = {2002},
  doi       = {10.1061/40650(2003)4}
}

@inproceedings{brown2020,
  author    = {Brown, Tom and Mann, Benjamin and Ryder, Nick and Subbiah, Melanie and Kaplan, Jared D. and Dhariwal, Prafulla and Neelakantan, Arvind and Shyam, Pranav and Sastry, Girish and Askell, Amanda and Agarwal, Sandhini and Herbert-Voss, Ariel and Krueger, Gretchen and Henighan, Tom and Child, Rewon and Ramesh, Aditya and Ziegler, Daniel and Wu, Jeffrey and Winter, Clemens and others},
  title     = {Language Models are Few-Shot Learners},
  booktitle = {Advances in Neural Information Processing Systems},
  volume    = {33},
  pages     = {1877--1901},
  year      = {2020}
}

@article{chandola2009,
  author  = {Chandola, Varun and Banerjee, Arindam and Kumar, Vipin},
  title   = {Anomaly detection: A survey},
  journal = {ACM Computing Surveys},
  volume  = {41},
  number  = {3},
  pages   = {15:1--15:58},
  year    = {2009},
  doi     = {10.1145/1541880.1541882}
}

@misc{equinor2018,
  author       = {{Equinor}},
  title        = {Disclosing all {Volve} data},
  year         = {2018},
  howpublished = {\url{https://www.equinor.com/news/archive/14jun2018-disclosing-volve-data}}
}

@article{gaddam2020,
  author  = {Gaddam, Anitha and Wilkin, Tim and Angelova, Maia and Gaddam, Jyothi},
  title   = {Detecting Sensor Faults, Anomalies and Outliers in the {Internet of Things}: A Survey on the Challenges and Solutions},
  journal = {Electronics},
  volume  = {9},
  number  = {3},
  pages   = {511},
  year    = {2020},
  doi     = {10.3390/electronics9030511}
}

@article{grimstad2026,
  author  = {Grimstad, Bjarne and Lundby, Eirik and Andersson, Henrik},
  title   = {{ManyWells}: Simulation of multiphase flow in thousands of wells},
  journal = {Geoenergy Science and Engineering},
  volume  = {257},
  pages   = {214226},
  year    = {2026},
  doi     = {10.1016/j.geoen.2025.214226}
}

@inproceedings{han2022,
  author    = {Han, Songqiao and Hu, Xiyang and Huang, Hailiang and Jiang, Minqi and Zhao, Yue},
  title     = {{ADBench}: Anomaly Detection Benchmark},
  booktitle = {Advances in Neural Information Processing Systems},
  volume    = {35},
  pages     = {32142--32159},
  year      = {2022}
}

@inproceedings{hegselmann2023,
  author    = {Hegselmann, Stefan and Buendia, Alejandro and Lang, Hunter and Agrawal, Monica and Jiang, Xiaoyi and Sontag, David},
  title     = {{TabLLM}: Few-shot Classification of Tabular Data with Large Language Models},
  booktitle = {Proceedings of The 26th International Conference on Artificial Intelligence and Statistics},
  pages     = {5549--5581},
  year      = {2023},
  url       = {https://proceedings.mlr.press/v206/hegselmann23a.html}
}

@article{hubert2018,
  author  = {Hubert, Mia and Debruyne, Michiel and Rousseeuw, Peter J.},
  title   = {Minimum covariance determinant and extensions},
  journal = {WIREs Computational Statistics},
  volume  = {10},
  number  = {3},
  pages   = {e1421},
  year    = {2018},
  doi     = {10.1002/wics.1421}
}

@misc{patel2025,
  author       = {Patel, Dhruvil and Lin, Shuyi and Rayfield, James and Zhou, Naiming and Vaculin, Roman and Martinez, Nestor and O'Donncha, Fearghal and Kalagnanam, Jayant},
  title        = {{AssetOpsBench}: Benchmarking {AI} Agents for Task Automation in Industrial Asset Operations and Maintenance},
  year         = {2025},
  howpublished = {arXiv preprint arXiv:2506.03828},
  doi          = {10.48550/arXiv.2506.03828}
}

@article{sandnes2021,
  author  = {Sandnes, Aleksander T. and Grimstad, Bjarne and Kolbj{\o}rnsen, Odd},
  title   = {Multi-task learning for virtual flow metering},
  journal = {Knowledge-Based Systems},
  volume  = {232},
  pages   = {107458},
  year    = {2021},
  doi     = {10.1016/j.knosys.2021.107458}
}

@article{souza2016,
  author  = {Souza, Francisco A. A. and Ara{\'u}jo, Rui and Mendes, Jorge},
  title   = {Review of soft sensor methods for regression applications},
  journal = {Chemometrics and Intelligent Laboratory Systems},
  volume  = {152},
  pages   = {69--79},
  year    = {2016},
  doi     = {10.1016/j.chemolab.2015.12.011}
}

@article{theyab2018,
  author  = {Theyab, Mohamed A.},
  title   = {Fluid Flow Assurance Issues: Literature Review},
  journal = {SciFed Journal of Petroleum},
  volume  = {2},
  number  = {1},
  pages   = {1--11},
  year    = {2018}
}

@inproceedings{tsai2024,
  author    = {Tsai, Chia-Peng and Teng, Grace and Wallis, Philip and Ding, Weiwei},
  title     = {{AnoLLM}: Large Language Models for Tabular Anomaly Detection},
  booktitle = {The Thirteenth International Conference on Learning Representations},
  year      = {2024},
  url       = {https://openreview.net/forum?id=7VkHffT5X2}
}

@book{tukey1977,
  author    = {Tukey, John W.},
  title     = {Exploratory Data Analysis},
  publisher = {Addison-Wesley},
  year      = {1977}
}

@article{vargas2019,
  author  = {Vargas, Ricardo Emmanuel Vaz and Munaro, Celso Jos{\'e} and Ciarelli, Patrick Marques and Medeiros, Andr{\'e} Gon{\c{c}}alves and Amaral, Bruno Guberfain Do and Barrionuevo, Daniel Celestino and Ara{\'u}jo, Jean Carlos Dias De and Ribeiro, Jorge Lu{\'i}s and Magalh{\~a}es, Lucas Pinheiro},
  title   = {A realistic and public dataset with rare undesirable real events in oil wells},
  journal = {Journal of Petroleum Science and Engineering},
  volume  = {181},
  pages   = {106223},
  year    = {2019},
  doi     = {10.1016/j.petrol.2019.106223}
}

@inproceedings{wu2026,
  author    = {Wu, Zhuo and Wang, Jingyi and Cheng, Zhen and Zhou, Yanan and Wang, Wenbin and Pu, Jian and Li, Chunwei and Ma, Chen},
  title     = {{ICAD-LLM}: One-for-All Anomaly Detection via In-Context Learning with Large Language Models},
  booktitle = {Proceedings of the AAAI Conference on Artificial Intelligence},
  volume    = {40},
  pages     = {15986--15994},
  year      = {2026},
  doi       = {10.1609/aaai.v40i19.38632}
}

@article{zha2025,
  author  = {Zha, Daochen and Bhat, Zaid Pervaiz and Lai, Kwei-Herng and Yang, Fan and Jiang, Zhimeng and Zhong, Shaochen and Hu, Xia},
  title   = {Data-centric Artificial Intelligence: A Survey},
  journal = {ACM Computing Surveys},
  volume  = {57},
  number  = {5},
  pages   = {129:1--129:42},
  year    = {2025},
  doi     = {10.1145/3711118}
}

@article{zhao2019,
  author  = {Zhao, Yue and Nasrullah, Zain and Li, Zheng},
  title   = {{PyOD}: A {Python} Toolbox for Scalable Outlier Detection},
  journal = {Journal of Machine Learning Research},
  volume  = {20},
  number  = {96},
  pages   = {1--7},
  year    = {2019}
}

\end{document}